\documentclass[sigconf, screen, nonacm]{acmart}
\AtBeginDocument{%
  }

\acmSubmissionID{XXXX}

\renewcommand\footnotetextcopyrightpermission[1]{}
\usepackage{soul}
\usepackage{url}
\usepackage{graphicx}
\usepackage{amsmath}
\usepackage{amsthm}
\usepackage{booktabs}
\usepackage{algorithm}
\usepackage{algorithmic}
\usepackage{bm}
\usepackage{color}
\usepackage{xcolor}
\usepackage{array}
\usepackage{colortbl}
\usepackage{multirow}
\usepackage{mathrsfs}
\usepackage{hhline}
\usepackage{mathtools}

\definecolor{codeblue}{RGB}{0, 82, 147}     
\definecolor{codegreen}{RGB}{0, 128, 0}    
\definecolor{codegray}{RGB}{100, 100, 100}  
\definecolor{codeorange}{RGB}{230, 145, 56} 
\definecolor{darkerblue}{rgb}{0,0.08,0.45}
\definecolor{royalblue}{RGB}{65,105,225}
\definecolor{lightblue}{RGB}{221,235,247}
\definecolor{figblue}{RGB}{47, 122, 232}  
\definecolor{figred}{RGB}{213, 32, 52}
\definecolor{figgreen}{RGB}{0, 137, 72} 
\definecolor{figyellow}{RGB}{217, 161, 5}
\definecolor{gray94}{gray}{.94}
\definecolor{gray90}{gray}{.90}

\newcommand{\blue}[1]{\textcolor{darkerblue}{#1}}
\definecolor{darkgreen}{RGB}{34,139,34}
\newcommand{\green}[1]{\textcolor{darkgreen}{#1}}
\newcommand{\gray}[1]{\textcolor{codegray}{#1}}
\newcommand{\gbf}[1]{\green{\bf{#1}}}

\definecolor{darkpurple}{RGB}{110,51,137}
\newcommand{\dpbf}[1]{\textcolor{darkpurple}{\bf{#1}}}

\newcolumntype{g}{>{\columncolor{gray94}}c} 
\newcommand{\grow}[1]{\rowcolor{gray94}{#1}} 
\newcommand{\brow}[1]{\rowcolor{lightblue}{#1}} 
\definecolor{lightgreen}{RGB}{212, 243, 234} 
\newcommand{\greenrow}[1]{\rowcolor{lightgreen}{#1}} 
\definecolor{lightorange}{RGB}{255, 239, 205} 
\usepackage{pifont}
\newcommand\graycross{\textcolor[rgb]{ .502,  .502,  .502}{\ding{55}}}

\makeatletter

\newcommand{\Rmnum}[1]{\expandafter\@slowromancap\romannumeral #1@}
\makeatother

\begin{document}

\title{Improving Radio Interferometry Imaging by Explicitly Modeling Cross-Domain Consistency in Reconstruction}


\author{Kai Cheng}
\affiliation{%
  \institution{The Hong Kong University of Science and Technology}
  \city{Hong Kong}
  \country{China}}
\email{kai.cheng@connect.ust.hk}

\author{Ruoqi Wang}
\affiliation{%
  \institution{The Hong Kong University of Science and Technology (Guangzhou)}
  \city{Guangzhou}
  \country{China}
}
\email{rwang280@connect.hkust-gz.edu.cn}

\author{Qiong Luo}
\authornote{Qiong Luo is the corresponding author.}
\affiliation{%
  \institution{The Hong Kong University of Science and Technology}
  \city{Hong Kong}
  \country{China}
}
\affiliation{%
  \institution{The Hong Kong University of Science and Technology (Guangzhou)}
  \city{Guangzhou}
  \country{China}
}
\email{luo@ust.hk}


\begin{abstract}
  Radio astronomy plays a crucial role in understanding the universe, particularly within the realm of non-thermal astrophysics. Images of celestial objects are derived from the signals (called visibility) measured by radio telescopes. Such imaging results, called dirty images, contain artifacts due to factors such as sparsity and therefore require reconstruction to improve imaging quality. Existing methods typically restrict reconstruction to a unimodal domain, either to the dirty image after imaging or to the sparse visibility prior to imaging. Focusing solely on each unimodal reconstruction results in the loss of complementary in-context information in either the visibility or image domain, leading to an incomplete modeling of mutual dependency and consistency. To address these challenges, we propose CDCRec, a multimodal radio interferometric data reconstruction method that explicitly models cross-domain consistency. We design a hierarchical multi-task and multi-stage framework to enhance the exploration of interplays between domains during reconstruction. Our experimental results demonstrate that CDCRec improves imaging performance through enhanced cross-domain correlation extraction. In particular, our self-supervised complementary modeling strategy is better than current methods at interferometric domain translations that rely heavily on recovering dense information from constrained source-domain data.
\end{abstract}

\ccsdesc[500]{Applied computing~Astronomy}

\keywords{Multimodal Learning, Radio Interferometry, Data Reconstruction}

\maketitle

\section{Introduction}
\begin{figure}[t]
  \centering
  \includegraphics[width=\linewidth]{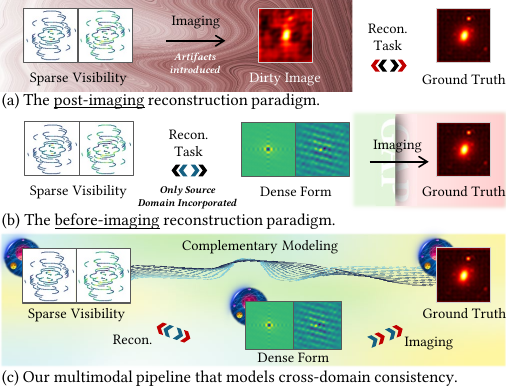}
  \caption{Comparison of the traditional reconstruction paradigm and our multimodal pipeline.}
  \label{fig:fig1}
\end{figure}

In radio astronomy, telescopes are deployed to observe the sky by recording radio waves and storing the distribution in the complex form of \textit{visibility} data, serving as sources for subsequent astronomical studies \cite{nimmo2025magnetospheric,li2026sudden}. Specifically, the cross-correlation between pairs of antenna signals is captured by radio interferometers in the frequency domain, holding astronomical significance. Compared to the visible spectrum, radio waves are able to penetrate dust and gas due to longer wavelengths, thus providing more comprehensive and insightful information. However, due to technical limitations of telescopes \cite{schmidt2022deep}, observed visibility data in the \textit{uv-plane} are spectrally sparse as certain regions cannot be sampled. Additionally, these data are affected by various factors such as cosmic microwave background radiation, galactic emissions, atmospheric disturbances, and receiver interferences \cite{wilson1979cosmic,singh2022detection}. Consequently, the \textit{dirty images} reconstructed from the information-incomplete \textit{sparse visibility} data severely hinder scientific research \cite{bouman2016computational,connor2022deep}.

Researchers have proposed to perform image reconstruction based on dirty images to remove some artifacts and noises \cite{cornwell2008multiscale,xie2022measurement}, as shown in Figure ~\ref{fig:fig1} (a). Also, recent studies \cite{wang2024polarrec,wang2025visrec} indicate that reconstructing visibility in the frequency domain yields improvements. The process is shown in Figure ~\ref{fig:fig1} (b). Fundamentally, each paradigm independently conducts reconstruction within a unimodal domain, hindering the exploration of mutual interactions and consistency between image and visibility. However, effectively establishing complementary modeling to connect both domains is critical for achieving a smooth domain transformation and obtaining high-fidelity \textit{clean images}. In this paper, we propose a multimodal radio interferometric data reconstruction approach that explicitly maintains cross-domain consistency to enhance imaging results.

The first challenge arises from the highly ill-posed nature of reconstructing the image from sparse visibility data. The inherent sparsity inevitably forms an underdetermined system \cite{ji2025recursive,beliaev2025inverse}. Such domain mapping is vulnerable to noise and easily degraded by massive artifacts, undermining its stability. Most existing methods adopt a straightforward yet insufficient target task of reconstruction, directly mapping the sparse visibility to its dense form and anticipating that such reconstructed results can be translated into precise images. However, this approach is prone to inadequate generalization due to the limited information available in sparse visibility. Specifically, insufficient semantic awareness hinders the ability of the network's convergence to distinguish multiple plausible predictions in blank spaces \cite{yun2025learning,nascimento2026towards}, making it fall short in addressing such ambiguity. This issue hinders the mode coverage of potential outcomes, thereby restricting the capacity to learn the diversity of the distribution. Relying solely on this straightforward task of reconstruction is unable to capture the essential semantic information, ultimately propagating the ambiguity into the reconstructed images.

The second challenge lies in the modeling of strong dependencies between the visibility and image domains. Both the intensity distribution and semantic structures of images are embedded within the frequency domain \cite{tatsunami2024fft,wang2025discovering}. Recovering key features of images from the underlying information in sparse visibility requires a joint modeling that leverages both domains. Performing unimodal image reconstruction after imaging inevitably confounds the domain transformation with noise, as unmitigated frequency-domain deficiency caused by sparsity propagates through the transformation \cite{cohen2018distribution}. Specifically, the missing knowledge about blank spaces introduces noise with uncertain patterns that are hard to trace in the image domain, making pattern recognition for removing such noise challenging. Furthermore, reconstructing visibility data before imaging makes the reconstruction and domain transformation separate. The absence of prior knowledge about the mutual dependency between domains renders the reconstruction limited.

To address these challenges, we propose CDCRec, a multimodal radio interferometric data reconstruction approach that explicitly models cross-domain consistency through a hierarchical two-stage learning strategy. CDCRec is designed to enhance the coherence between reconstruction and imaging processes, underscoring explicit modeling of dependencies between domains, as illustrated in Figure ~\ref{fig:fig1} (c). To strengthen the joint learning of multimodal representations, we develop two stages that progressively extract highly compressed information and preserve critical information across domains. Compared with previous unimodal methods, CDCRec focuses on cross-domain alignment and representation learning between visibility data and images by explicitly incorporating consistency constraints. Building upon the learned priors, our interferometric data reconstruction is effective and produces high-fidelity imaging results.

Specifically, we incorporate contrastive learning and complementary masking reconstruction as pretext tasks to explicitly enable the model to learn cross-domain knowledge extraction. Since the problematic behavior of learning the shortcut for averaging plausible predictions of distinct samples is caused by the sparsity of visibility, CDCRec penalizes it through the regularization of the contrastive loss and enforces the model to establish proper pairings between visibility data and corresponding images. To make the learned representations more compact, we design an additional pretext task of complementary masking reconstruction to further strengthen the extractions of multimodal dual-domain interplay, thereby enhancing the confidence of the target task of interferometric data reconstruction. To further promote diversity, we propose to apply random masking by performing band limiting to visibility and patchification to the image based on the characteristics of both modalities. Through modeling cross-domain consistency, CDCRec learns multimodal features to address the limitations and improve the effectiveness of interferometric data reconstruction.

Overall, our contributions can be summarized as follows:

\begin{itemize}
\item We propose a multimodal radio interferometric data reconstruction approach, called CDCRec, explicitly modeling cross-domain consistency and therefore addressing the challenges of sparse visibility.
\item We propose a hierarchical framework with effective strategies, which seamlessly integrates two pretext tasks to enable multimodal learning from mutually dependent domains to replace the previous paradigm of a single-target task with insufficient modeling ability.
\item We empirically show that our CDCRec improves the accuracy and confidence of imaging after the enhancement of the modeling of cross-domain consistency. Extended experimental results show the effectiveness and robustness of our proposed method.
\end{itemize}

\section{Background and Related Work}
\subsection{Radio Interferometric Imaging}
In modern astrophysics, the extension of the observed spectral bands toward the electromagnetic spectrum has greatly enriched our understanding of the state and evolutionary history of cosmic matter. Radio astronomy is able to perform observations at extremely low temperatures, while optical astronomy is primarily limited to high temperatures \cite{xu20220,chime2022sub}. Meanwhile, with long wavelengths and strong penetration capabilities, radio waves enable the discovery of densely obscured astrophysical environments \cite{steyn2024h}. Spatial resolution was a major limiting factor in the early development of radio astronomy. According to the Rayleigh Criterion \cite{rayleigh1880investigations}, achieving resolutions comparable to optical telescopes with single-aperture radio telescopes requires extremely large physical apertures. Therefore, Very Long Baseline Interferometry (VLBI) \cite{bouman2018reconstructing} is implemented by using separated antenna arrays for spatial frequency sampling. However, the number of baselines remains limited, leading to essentially sparse observations due to the discrete samplings in the uv-plane \cite{thompson2017interferometry,bouman2018reconstructing} and making reconstruction indispensable.

According to the van Cittert-Zernike theorem \cite{zernike1938concept,goodman2015statistical}, the complex spatial mutual coherence function measured in the observation plane, namely visibility, is strictly equal to the two-dimensional Fourier transform of the surface brightness distribution function $I(l, m)$ of a distant celestial body, given by:
\begin{equation}
  I(l, m) = \int_u \int_v \bar{\mathcal{M}}_v \left(u,v\right) e^{j 2\pi (ul + vm)} \, du \, dv,
\end{equation}
where $\bar{\mathcal{M}}_v \left(u,v\right)$ denotes the frequency-domain visibility data and $I(l, m)$ denotes the intensity distribution of the sky.

\subsection{Interferometric Data Reconstruction}
Most existing methods restricted reconstruction to a unimodal domain. On the one hand, some methods performed reconstruction in the spatial domain after imaging. Specifically, the CLEAN algorithm \cite{cornwell2008multiscale} iteratively removed the dirty beam from dirty images, suppressing coherent interference. Although its assumption of point-like sources \cite{connor2022deep} simplified the problem mathematically, it performed poorly on sources with complex structures, resulting in its fundamental limitation. It focused solely on reconstructing the spatial domain and failed to capture the essential information preserved in the spectral domain. Other methods made efforts to utilize dirty images and convert them into clean images with the help of deep learning-based approaches, which are later demonstrated to be less effective due to the redundant noise before imaging \cite{bouman2016computational,connor2022deep}. 

On the other hand, some methods performed reconstruction before imaging, which are typically limited to the spectral domain. Specifically, Schmidt \textit{et al.} \cite{schmidt2022deep} and Wu \textit{et al.} \cite{wu2022neural} introduced convolutional neural networks and sequence models for reconstructing incomplete data, respectively. Wang \textit{et al.} \cite{wang2024polarrec} proposed to consider all components in the angular coordinates of visibility data. Wang \textit{et al.} \cite{wang2025visrec} also explored the data augmentation in the spectral domain. In comparison to existing work, our CDCRec proposes the multimodal modeling of the cross-domain consistency and improves the imaging results by explicitly learning their multimodal interplay and dependency.

\section{Methodology}
\subsection{Overall Architecture}
\begin{figure*}[t]
  \centering
  \includegraphics[width=\linewidth]{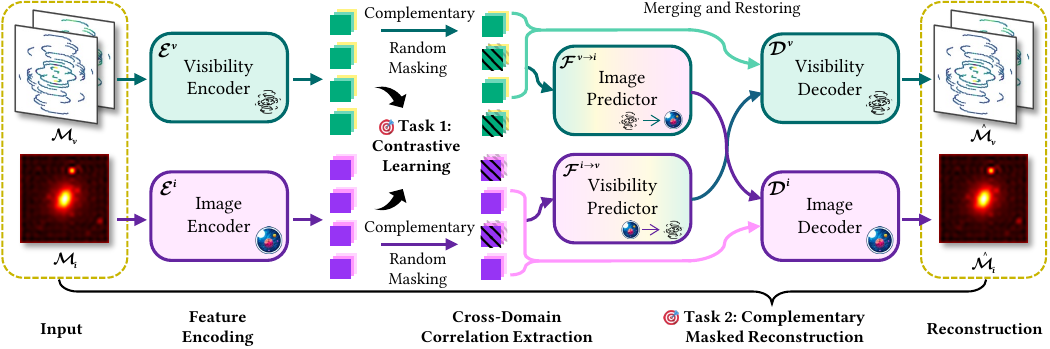}
  \caption{Overview of the stage of Self-supervised Complementary Modeling (SCM). The visibility encoder $\mathcal{E}^v$ and image encoder $\mathcal{E}^i$ take $\mathcal{M}_v$ and $\mathcal{M}_i$ as input and output encoded features $\xi$ and $\eta$ in the latent space, respectively. The first pretext task helps align the encoded latent space based on contrastive learning. After performing complementary masking to $\xi$ and $\eta$ following the principles of the band limiting and image patchification, the image predictor $\mathcal{F}^{v \rightarrow i}$ and visibility predictor $\mathcal{F}^{i \rightarrow v}$ predict the masked parts of each other. Finally, visible and predicted parts are merged and decoded by visibility decoder $\mathcal{D}^v$ and image decoder $\mathcal{D}^i$ accordingly to complete the second pretext task of complementary masking reconstruction and output the results of $\hat{\mathcal{M}}_v$ and $\hat{\mathcal{M}}_i$.}
  \label{fig:fig2}
\end{figure*}

\begin{figure}[t]
  \centering
  \includegraphics[width=\linewidth]{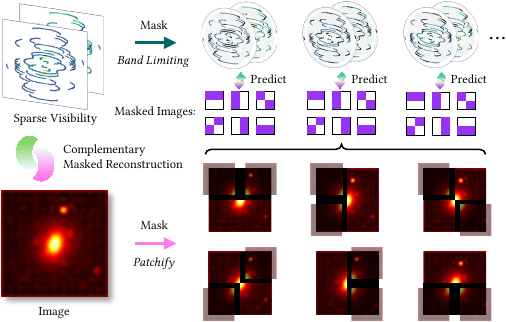}
  \caption{Illustration of the complementary masking in CDCRec. The sparse visibility is masked by applying band limiting, while the image is first patchified and then masked. The masked parts are depicted in black. Due to sparsity, we adopt random sampling within each band to ensure the balance between the masked and visible information.
  }
  \label{fig:fig3}
\end{figure}

The overall framework of the first stage of Self-supervised Complementary Modeling (SCM) in CDCRec is depicted in Figure ~\ref{fig:fig2}. Sparse visibility $\mathcal{M}_v \in \tilde{\mathcal{S}}_{v}$ and its corresponding image $\mathcal{M}_i \in \mathcal{S}_{i}$ are input to explore their cross-domain consistency. Specifically, $\mathcal{M}_v$ can be represented through applying the Fourier Transformation $\mathcal{F}$ to $\mathcal{M}_i$ with an under-sampling matrix $\bm{D}_{\Lambda}$ and noise $\epsilon$, thus making the inverse problem ill-posed, which is defined as:
\begin{equation}
\begin{split}
    \mathcal{M}^{v} = \bm{D}_{\Lambda} \mathcal{F}\left(\mathcal{M}_i\right) + \epsilon,
\end{split}
\end{equation}
where $\Lambda$ denotes the telescopes' sampling pattern. Thus, the ideal imaging process expects to perform Inverse Fourier Transformation $\mathcal{F}^{-1}$ to dense visibility $\bar{\mathcal{M}}_v=\mathcal{F}\left(\mathcal{M}_i\right) \in \bar{\mathcal{S}}_{v}$ without being affected by sparsity. This domain transformation can be expressed as:
\begin{equation}
    \bar{\mathcal{S}}_{v} \xmapsto{\mathcal{F}^{-1}} \mathcal{S}_{i}.
\end{equation}
However, radio telescopes can only acquire sparse visibility data, making the real scenario formulated as:
\begin{equation}
    \tilde{\mathcal{S}}_{v} \subset \bar{\mathcal{S}}_{v} \xmapsto{\mathcal{F}^{-1}} \tilde{\mathcal{S}}_{i},
\end{equation}
\begin{equation}
    p_{\tilde{\mathcal{S}}_{v}} = \int_{\bar{\mathcal{S}}_{v}} p_{\Lambda}\left( \mathcal{M}^{v} \vert x \right) p_{\bar{\mathcal{S}}_{v}}\left( x \right) dx,
\end{equation}
where $\tilde{\mathcal{S}}_{v}$ denotes the domain of sparse visibility, $p_{\tilde{\mathcal{S}}_{v}}$ is the representation of $\tilde{\mathcal{S}}_{v}$. Thus, there remains a large discrepancy between $\tilde{\mathcal{S}}_{i}$ and the desired $\mathcal{S}_{i}$, which can be measured by KL divergence $D_{KL}\left( \mathcal{S}_{i} \Vert  \tilde{\mathcal{S}}_{i} \right)$. One solution to this problem is to reconstruct sparse visibility into its dense form. As such, the reconstruction quality will have an indispensable effect on the final results. This domain transformation can be formulated as:
\begin{equation}  \label{eq:other_trans}
    \underbrace{\tilde{\mathcal{S}}_{v} \xmapsto{\quad \mathcal{E}_\theta \quad} \hat{\mathcal{S}}_{v}}_{\substack{\text{Visibility Domain} \\ \text{Transformation Term}}} \xmapsto{\mathcal{F}^{-1}} \hat{\mathcal{S}}_{i},
\end{equation}
where $\hat{\mathcal{S}}_{v}$ can be viewed as an approximation of $\bar{\mathcal{S}}_{v}$ and thus the final transformation result $\hat{\mathcal{S}}_{i}$ is also an approximation of $\mathcal{S}_{i}$, making the objective of model $\mathcal{E}_\theta$ the minimization of $D_{KL}\left( \mathcal{S}_{i} \Vert \hat{\mathcal{S}}_{i} \right)$. In Equation \eqref{eq:other_trans}, we can observe that the reconstruction process is restricted to only the visibility domain. Such a visibility domain transformation term falls short in learning the cross-domain consistency, weakening the utilization of dependencies between source and target domains in the subsequent domain transformation. Thus, we propose to first explore the mutual interactions between the visibility and the image. Data utilized in the stage of SCM are sampled from $\tilde{\mathcal{S}}_{v}$ and $\mathcal{S}_{i}$ and formulated as pairs $\mathcal{D}=\left\{ \mathcal{M}_{v,k}, \mathcal{M}_{i,k} \right\}^{m}_{k=1}$, where $\mathcal{M}_{v,k} \in \tilde{\mathcal{S}}_{v}$ and $\mathcal{M}_{i,k} \in \mathcal{D}_{i}=\left\{ \left( \mathcal{M}_{i} \right) \in \mathcal{S}_{i} \right\}$. The domain transformation of our CDCRec can be expressed as:
\begin{equation}
    \mathcal{D}_{i} + \tilde{\mathcal{S}}_{v} \xmapsto{p^{CDCRec}_{\theta}} \hat{\mathcal{S}}_{v} \xmapsto{\mathcal{F}^{-1}} \hat{\mathcal{S}}_{i} ,
\end{equation}
\begin{equation}
    \min D_{KL}\left( \mathcal{S}_{i} \Vert \hat{\mathcal{S}}_{i} \right),
\end{equation}
where $p^{CDCRec}_{\theta}$ is the representation of the CDCRec model.

\begin{figure}[t]
  \centering
  \includegraphics[width=\linewidth]{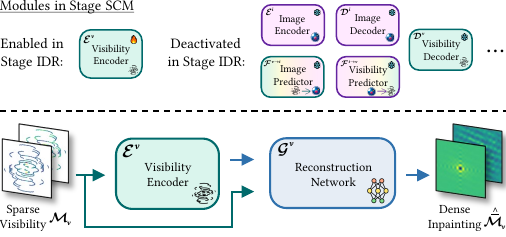}
  \caption{Illustration of the stage of Interferometric Data Reconstruction (IDR).
  }
  \label{fig:fig4}
\end{figure}

Next, we provide essential definitions and introductions of components in our framework to facilitate subsequent derivations. In the stage of SCM, the visibility encoder $\mathcal{E}^v$ and image encoder $\mathcal{E}^i$ take $\mathcal{M}_v$ and $\mathcal{M}_i$ as input and output encoded features $\xi$ and $\eta$ in the latent space, respectively, as shown in Figure ~\ref{fig:fig2}. We design the first pretext task based on contrastive learning to align two modalities in the encoded latent space. We perform complementary masking to $\xi$ and $\eta$ following the principles of the band limiting and image patchification, where the detail is shown in Figure ~\ref{fig:fig3}. Next, the image predictor $\mathcal{F}^{v \rightarrow i}$ and visibility predictor $\mathcal{F}^{i \rightarrow v}$ predict masked parts of each modality conditioned on their respective visible components of their counterparts. Finally, we merge the visible and predicted parts and input them into visibility decoder $\mathcal{D}^v$ and image decoder $\mathcal{D}^i$ accordingly, outputing the results of $\hat{\mathcal{M}}_v$ and $\hat{\mathcal{M}}_i$ and completing the second pretext task of complementary masking reconstruction. In the stage of Interferometric Data Reconstruction (IDR), only the visibility encoder $\mathcal{E}^v$ from the stage of SCM is retained and fine-tuned, as shown in Figure ~\ref{fig:fig4}. Furthermore, we design a reconstruction network based on Multilayer Perceptron (MLP), which takes $\hat{\mathcal{M}}_v$ and $\xi$ as input and outputs the dense form of visibility $\hat{\bar{\mathcal{M}}}_v$.

Building upon the preliminary knowledge introduced above, we proceed to derive the overall objective function, given by: 
\begin{equation}  \label{eq:overallobj1}
    \resizebox{.91\columnwidth}{!}{$
    \begin{aligned}
        \underset{\theta}{\arg \min} \, D_{KL} \left( \mathcal{S}_{i} \Vert \hat{\mathcal{S}}_{i}  \right)  &= \underset{\theta}{\arg \min} \, -\int_{\mathcal{M}_{i}} p_{\mathcal{S}_{i}}\left( \mathcal{M}_{i} \right) \log \frac{p_{\hat{\mathcal{S}}_{i}}\left( \mathcal{M}_{i} \right)}{p_{\mathcal{S}_{i}}\left( \mathcal{M}_{i} \right)}  \,d\mathcal{M}_{i}  \\
        &= \underset{\theta}{\arg \max} \, \int_{\mathcal{M}_{i}} p_{\mathcal{S}_{i}}\left( \mathcal{M}_{i} \right) \log p_{\hat{\mathcal{S}}_{i}}\left( \mathcal{M}_{i} \right)\\
        &- \underset{\theta}{\arg \max} \, \int_{\mathcal{M}_{i}} p_{\mathcal{S}_{i}}\left( \mathcal{M}_{i} \right) \log p_{\mathcal{S}_{i}}\left( \mathcal{M}_{i} \right)  \,d\mathcal{M}_{i}\\
    \end{aligned}
    $}.
\end{equation}
Since the second term in Equation \eqref{eq:overallobj1} is independent of the parameters $\theta$ being optimized, it can be disregarded. Thus, Equation \eqref{eq:overallobj1} simplifies to:
\begin{equation}  \label{eq:overallobj2}
    \begin{aligned}
        \underset{\theta}{\arg \min} \, D_{KL} \left( \mathcal{S}_{i} \Vert \hat{\mathcal{S}}_{i}  \right)  &= \underset{\theta}{\arg \max} \, \int_{\mathcal{M}_{i}} p_{\mathcal{S}_{i}}\left( \mathcal{M}_{i} \right) \log p_{\hat{\mathcal{S}}_{i}}\left( \mathcal{M}_{i} \right)\\
        &= \underset{\theta}{\arg \max} \, \mathbb{E}_{\mathcal{M}_{i} \sim p_{\mathcal{S}_{i}}} \left[ \log p_{\hat{\mathcal{S}}_{i}}\left( \mathcal{M}_{i} \right) \right] \\
        &= \underset{\theta}{\arg \max} \,\log  \prod_{k = 1}^{m} \mathcal{F}^{-1} \left( p_{\hat{\mathcal{S}}_{v}}\left( \hat{\mathcal{M}}_{v,k} \right) \right)  \\
    \end{aligned}.
\end{equation}
The distribution of the modeling of CDCRec can be expressed as:
\begin{equation}  \label{eq:svtilde}
    \hat{\mathcal{S}_{v}} \sim p_{\hat{\mathcal{S}}_{v}} = p^{CDCRec}_\theta = p_{\theta, \mathcal{M}^v \sim \tilde{\mathcal{S}}_v, \mathcal{M}^i \sim \mathcal{D}_i} \left( \mathcal{M}^v, \mathcal{M}^i \right) .
\end{equation}
Substituting Equation \eqref{eq:svtilde} into Equation \eqref{eq:overallobj2}, we obtain:
\begin{equation}  \label{eq:overallobj3}
    \resizebox{.89\columnwidth}{!}{$
    \begin{aligned}
        \underset{\theta}{\arg \min} \, D_{KL} \left( \mathcal{S}_{i} \Vert \hat{\mathcal{S}}_{i}  \right)  &= \underset{\theta}{\arg \max} \, p_{\theta, \mathcal{M}_v \sim \tilde{\mathcal{S}}_v, \mathcal{M}_i \sim \mathcal{D}_i}\left( \mathcal{M}_{v}, \mathcal{M}_{i} \right)  \\
    \end{aligned}
    $}.
\end{equation}
In summary, the objective of enhancing imaging quality is rendered equivalent to the optimization of our proposed CDCRec. In the following, we detail our method in a stage-wise manner.

\subsection{Stage of SCM}
\subsubsection{Contrastive Learning.} 

In the stage of SCM, our main purpose is to grasp the cross-domain consistency with the aid of two pretext tasks. Thus, the visibility encoder $\mathcal{E}^v$ and image encoder $\mathcal{E}^i$ should not only embed the two modalities into latent spaces, but also learn to establish connections between each other in a unified space. The encoding processes can be expressed as:
\begin{equation}  \label{eq:visencoder}
    \xi = \mathcal{E}^{v} (\mathcal{M}_{v}) \sim p^{\mathcal{E}^{v}} \left( \xi \vert \mathcal{M}_{v} \right) ,
\end{equation}
\begin{equation}
    \eta = \mathcal{E}^{i} (\mathcal{M}_{i}) \sim p^{\mathcal{E}^{i}} \left( \eta \vert \mathcal{M}_{i} \right) .
\end{equation}
We perform the mutual information maximization between modalities to achieve alignment, defined as $\mathcal{I} \left( \xi; \eta \right)$. Also, the joint representation of $\xi$ and $\eta$ is denoted as:
\begin{equation}
    p\left(\xi,\eta\right) = \mathbb{E}_{\mathcal{M}_v,\mathcal{M}_i} \left[ p^{\mathcal{E}^{v}} \left( \xi \vert \mathcal{M}_{v} \right) p^{\mathcal{E}^{i}} \left( \eta \vert \mathcal{M}_{i} \right) \right] .
\end{equation}
According to Oord \textit{et al.} \cite{oord2018representation}, minimizing the information noise-contrastive estimation loss is equivalent to maximizing the lower bound of $\mathcal{I} \left( \xi; \eta \right)$. Thus, we construct all possible bi-modal pairs within a mini-batch composed of $n$ instances, where $\left( \xi_j,\eta_j \right) \sim p\left( \xi,\eta \right), \, j=1,\dots,n$ represent positive pairs while others represent negative pairs. After transforming the density ratio estimation into classification, we minimize the negative lower bound as follows:
\begin{equation}
    \mathcal{L}^{v \rightarrow i} = - \mathbb{E}_{p^{\mathcal{E}^{v}},p^{\mathcal{E}^{i}}} \left[ \log {\frac{\exp \left(\left\Vert{\xi_{l}}\right\Vert^T \left\Vert{\eta_{l}}\right\Vert \right) /\tau}{\sum\nolimits_{j = 1}^n \left( \exp \left(\left\Vert{\xi_{l}}\right\Vert^T \left\Vert{\eta_{j}}\right\Vert \right) /\tau \right)}}\right] ,
\end{equation}
where $\tau$ denotes the temperature. $\mathcal{L}^{v \rightarrow i}$ represents the loss for identifying the positive $\eta$ given $\xi$, and vice versa for $\mathcal{L}^{i \rightarrow v}$. Thus, the total contrastive loss is defined as:
\begin{equation}
    \mathcal{L}^{c} = \frac{1}{2} \left( \mathcal{L}^{v \rightarrow i} + \mathcal{L}^{i \rightarrow v} \right).
\end{equation}

\begin{figure*}[t]
  \centering
  \includegraphics[width=\linewidth]{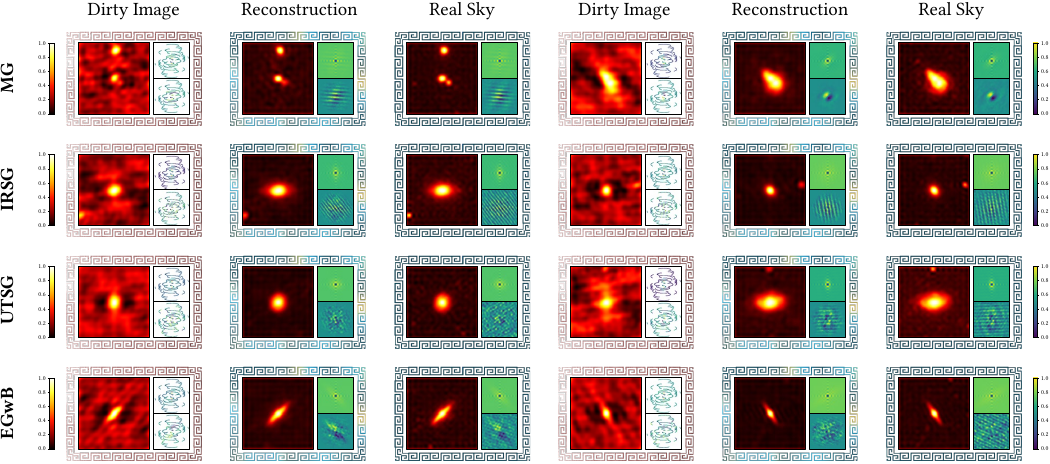}
  \caption{Visualization of the reconstruction results of CDCRec compared with the dirty image and real sky, including both visibility and image. In each panel, the upper half of the visibility is the real part, and the lower half is the imaginary part.}
  \label{fig:fig5}
\end{figure*}

\subsubsection{Complementary Masking Reconstruction.} 

The second pretext task further strengthens the alignment and exploits cross-domain consistency. The masking on two modalities is denoted as:
\begin{equation}
    \xi^{visible} = \mathcal{B} (\xi) \odot \Omega ,
\end{equation}
\begin{equation}
    \eta^{visible} = \mathcal{P} (\eta) \odot \left( 1-\Omega \right) ,
\end{equation}
where $\Omega$ is the masking matrix, $\mathcal{B}$ is the patchifying operation for visibility based on band limiting, and $\mathcal{P}$ is the operation for image patchification. Then, the visible parts are utilized to mutually predict the complementary masked parts, which is denoted as:
\begin{equation}
    \hat{\eta}^{masked} = \mathcal{F}^{v \rightarrow i} (\xi^{visible}) \sim p^{v \rightarrow i} \left( \hat{\eta}^{masked} \, \middle| \, \xi^{visible} \right) ,
\end{equation}
\begin{equation}
    \hat{\xi}^{masked} = \mathcal{F}^{i \rightarrow v} (\eta^{visible}) \sim p^{i \rightarrow v} \left( \hat{\xi}^{masked} \, \middle| \, \eta^{visible} \right) ,
\end{equation}
where $\hat{\xi}^{masked}$ and $\hat{\eta}^{masked}$ are the predictions. Then, the combinations of visible and predicted masked parts are denoted as:
\begin{equation}
    \hat{\xi} = \xi^{visible} \odot \Omega + \hat{\xi}^{masked} \odot \left( 1-\Omega \right) ,
\end{equation}
\begin{equation}
    \hat{\eta} = \eta^{visible} \odot \left( 1-\Omega \right) + \hat{\eta}^{masked} \odot \Omega .
\end{equation}
Next, $\mathcal{D}^v$ and $\mathcal{D}^i$ map the latent representations to output the complementary reconstructions $\hat{\mathcal{M}}_v$ and $\hat{\mathcal{M}}_i$, denoted as:
\begin{equation}
    \hat{\mathcal{M}}_v = \mathcal{D}^v \left( \hat{\xi} \right) \sim p^{\mathcal{D}^v} \left( \hat{\mathcal{M}}_v \, \middle| \, \hat{\xi} \right) ,
\end{equation}
\begin{equation}
    \hat{\mathcal{M}}_i = \mathcal{D}^i \left( \hat{\eta} \right) \sim p^{\mathcal{D}^i} \left( \hat{\mathcal{M}}_i \, \middle| \, \hat{\eta} \right) .
\end{equation}
For the reconstruction of $\mathcal{M}_v$, our objective is to maximize the log-likelihood $\log p \left( \mathcal{M}_v \right)$. By applying Jensen's inequality, we can derive the Evidence Lower Bound (ELBO) as follows:
\begin{equation}
    \begin{aligned}
        \log p \left( \mathcal{M}_v \right)  &= \log \int p \left( \mathcal{M}_v, \xi \right) d\xi\\
        &= \log \mathbb{E}_{\xi \sim p^{\mathcal{E}^v}} \left[ \frac{p \left( \mathcal{M}_v, \xi \right)}{p^{\mathcal{E}^v} \left( \xi \middle| \mathcal{M}_v \right)} \right] \\
        &\geq \mathbb{E}_{\xi \sim p^{\mathcal{E}^v}} \left[ \log \frac{p \left( \mathcal{M}_v, \xi \right)}{p^{\mathcal{E}^v} \left( \xi \middle| \mathcal{M}_v \right)} \right] = \text{ELBO}^v \\
    \end{aligned},
\end{equation}
\begin{equation}  \label{eq:elbo}
    \resizebox{.89\columnwidth}{!}{$
    \begin{aligned}
        \text{ELBO}^v  &= \mathbb{E}_{\xi \sim p^{\mathcal{E}^v}} \left[ \log \frac{p^{\mathcal{D}^v} \left( \mathcal{M}_v \middle| \xi \right) p \left( \xi \right)}{p^{\mathcal{E}^v} \left( \xi \middle| \mathcal{M}_v \right)} \right] \\
        &= \mathbb{E}_{\xi \sim p^{\mathcal{E}^v}} \left[ \log p^{\mathcal{D}^v} \left( \mathcal{M}_v \middle| \xi \right) \right]   -   \mathcal{D}_{KL} \left( p^{\mathcal{E}^v} \left( \xi \middle| \mathcal{M}_v \right) \, \middle\| \, p \left( \xi \right) \right)    \\
    \end{aligned}
    $}.
\end{equation}
Since the latter term in Equation \eqref{eq:elbo} also constrains the latent representations, we choose to streamline the objective and consider only the former term to avoid redundancy and potential conflicts with $\mathcal{L}^{c}$. Thus, we quantify the reconstruction loss using the $L_2$ distance between $\mathcal{M}_v$ and $\hat{\mathcal{M}}_v$:
\begin{equation}
    \mathcal{L}^{rec}_v = \left\Vert \mathcal{M}_v - \hat{\mathcal{M}}_v \right\Vert_2^2.
\end{equation}
Similarly, we have the reconstruction loss of the image as follows:
\begin{equation}
    \mathcal{L}^{rec}_i = \left\Vert \mathcal{M}_i - \hat{\mathcal{M}}_i \right\Vert_2^2.
\end{equation}
In summary, the overall loss in the stage of SCM is composed of $\mathcal{L}^{rec}_v$, $\mathcal{L}^{rec}_i$, and $\mathcal{L}^c$ and balanced by a hyperparameter called contrastive loss weight $\kappa$, which is defined as follows:
\begin{equation}
    \mathcal{L}^{SCM} = \mathcal{L}^{rec}_v + \mathcal{L}^{rec}_i + \kappa \mathcal{L}^c.
\end{equation}

\subsection{Stage of IDR}
In the stage of IDR, the visibility encoder $\mathcal{E}^v$ from the stage of SCM is retained and fine-tuned. From Equation \eqref{eq:visencoder}, we obtain:
\begin{equation}
    \hat{\bar{\mathcal{M}}}_v = \mathcal{G}^{v} \left( \mathcal{E}^{v} \left(\mathcal{M}_{v} \right), \mathcal{M}_{v} \right) \sim p^{\mathcal{G}^{v}} \left( \hat{\bar{\mathcal{M}}}_v \middle\vert \, \xi, \mathcal{M}_{v} \right) ,
\end{equation}
where $\mathcal{G}^{v}$ denotes the reconstruction network. Following the work by Jiang \textit{et al.} \cite{jiang2021focal}, we optimize parameters in the stage of IDR by utilizing losses computed from complex values in the spectral domain:
\begin{equation}
    \begin{aligned}
        \mathcal{L}^{IDR} \left( \hat{\bar{\mathcal{M}}}_v, \bar{\mathcal{M}}_v \right) = \operatorname{Avg} \left( {\omega \left|\hat{\bar{\mathcal{M}}}_v - \bar{\mathcal{M}}_v\right|^2} \right)
    \end{aligned},
\end{equation}

\begin{equation}
    \begin{aligned}
        \omega = \left( \frac{\rho}{\max\left( \rho \right)} + 1 \right) \left|\hat{\bar{\mathcal{M}}}_v - \bar{\mathcal{M}}_v\right|
    \end{aligned},
\end{equation}
where $\operatorname{Avg}$ denotes the average operation on the uv-plane and $\rho$ denotes the amplitude.

\section{Experiments}

\subsection{Experimental Setup}
Following Wang \textit{et al.} \cite{wang2024polarrec}, we utilize identical telescope configurations to sample interferometric data using the eht-imaging toolkit \cite{chael2018interferometric,chael2019ehtim} from real astronomical observations in different public datasets of distinct galaxy morphologies: Merging Galaxies (MG), In-between Round Smooth Galaxies (IRSG), Unbarred Tight Spiral Galaxies (UTSG), and Edge-on Galaxies with Bulge (EGB). These data are sourced from the DESI Legacy Imaging Surveys \cite{dey2019overview}, integrating contributions from the Beijing-Arizona Sky Survey (BASS) \cite{zou2017project}, the DECam Legacy Survey (DE-CaLS) \cite{blum2016decam}, and the Mayall z-band Legacy Survey \cite{silva2016mayall}. To ensure a fair comparison with previous baseline models, we use the same observation parameters as in the 8-telescope Event Horizon Telescope (EHT). All proposed models and experiments are implemented in PyTorch \cite{paszke2019pytorch}. The training and evaluation processes are conducted on a high-performance Linux workstation equipped with an NVIDIA H800 GPU. To quantitatively analyze the quality of reconstructions, we adopt two commonly used metrics, Peak Signal-to-Noise Ratio (PSNR) and Structural Similarity Index Measure (SSIM), for evaluating the generated images.

\subsection{Overall Performances}
Table ~\ref{tab:table1} presents the quantitative comparison of our proposed CDCRec against previous methods on four datasets, where the best and second-best results are highlighted in bold and underlined, respectively. As shown in the table, our multimodal method consistently outperforms competing approaches across both metrics. Specifically, our CDCRec yields an average improvement of 1.99 over the second-best methods in terms of PSNR. Compared to methods that reconstruct the image in a unimodal setting, the improvement is more significant, reaching 8.28. These performance gains demonstrate that our method suppresses artifacts and noise effectively throughout the reconstruction and imaging process. Regarding structural preservation, improvements in SSIM further demonstrate the enhanced ability of CDCRec to retain details, which are crucial for astronomical observations. Such performance improvements showcase that explicit modeling of cross-domain consistency contributes key information for producing more precise images. 

We further present a visual comparison of the dirty image, reconstruction result generated by CDCRec, and real sky across different datasets and their corresponding visibilities, as shown in Figure ~\ref{fig:fig5}. In contrast to dirty images that suffer from severe artifacts, CDCRec effectively produces reconstructions that are more consistent with the real sky, better recovering complex details, such as edges and textures. This improvement shows that multimodal features extracted from both domains carry mutual dependencies for domain transformations and preserve crucial features to facilitate a highly intact reconstruction. Overall, the qualitative visualization results demonstrate the effectiveness of enhancing the consistency between visibility and image, providing better support for subsequent scientific studies \cite{wang2025galaxalign}.

\begin{table*}[t]
    \centering
    \caption{Performance comparisons on different datasets. The Peak Signal to Noise Ratio (PSNR$\uparrow$: higher is better) and Structural Similarity Index Measure (SSIM$\uparrow$) are reported. \textbf{Bold} and \underline{underlined} denote the best and second best results respectively.}
    \begin{tabular}{lc|cccccccc}
    \toprule
\multirow{2}{*}{\textbf{Method}}              & \multirow{2}{*}{\textbf{Recon. Modality}}      & \multicolumn{2}{c}{\textbf{MG}} & \multicolumn{2}{c}{\textbf{IRSG}} & \multicolumn{2}{c}{\textbf{UTSG}} & \multicolumn{2}{c}{\textbf{EGB}} \\
                             &                     & PSNR$\uparrow$            & SSIM$\uparrow$            & PSNR$\uparrow$             & SSIM$\uparrow$            & PSNR$\uparrow$             & SSIM$\uparrow$            & PSNR$\uparrow$            & SSIM$\uparrow$           \\ \cmidrule{1-10}
\grow DIRTY \cite{gilbert2014recent}           & \gray{N/A}    & 11.633          & 0.715           & 11.398           & 0.722           & 11.890           & 0.725           & 11.356          & 0.708          \\
CLEAN \cite{cornwell2008multiscale}                        & \multirow{2}{*}{\textit{Image (Unimodal)}}           & 18.557          & 0.818           & 20.913           & 0.839           & 17.048           & 0.806           & 19.448          & 0.831          \\
U-Net \cite{xie2022measurement}                      &            & 18.126          & 0.818           & 19.770           & 0.828           & 14.877           & 0.786           & 19.232          & 0.822          \\ \cmidrule{1-10}
Radionets \cite{schmidt2022deep}                 & \multirow{4}{*}{\textit{Visibility (Unimodal)}}            & 19.687          & 0.836           & 21.369           & 0.854           & 20.828           & 0.844           & 20.322          & 0.836          \\
NF \cite{wu2022neural}                  &   & 20.560          & 0.875           & 24.099           & 0.898           & 21.323           & 0.880           & 21.276          & 0.879        \\
VisRec \cite{wang2025visrec}                   &   & 22.267          & 0.883           & 24.940           & \underline{0.915}           & 23.805           & \underline{0.909}           & 23.268          & 0.901        \\
PolarRec \cite{wang2024polarrec}                        &   & \underline{24.088}     & \underline{0.884}           & \underline{26.234}      & 0.912           & \underline{25.384}      & 0.904           & \underline{25.420}     & \underline{0.908}          \\
\brow CDCRec(\textbf{Ours})          & \blue{\textit{Visibility (Multimodal)}}           & \gbf{26.506}          & \dpbf{0.914}           & \gbf{28.081}           & \dpbf{0.925}           & \gbf{27.497}           & \dpbf{0.924}           & \gbf{26.989}          & \dpbf{0.917}          \\
    \bottomrule
    \end{tabular}
    \label{tab:table1}
\end{table*}

\begin{figure}[t]
  \centering
  \includegraphics[width=.8\linewidth]{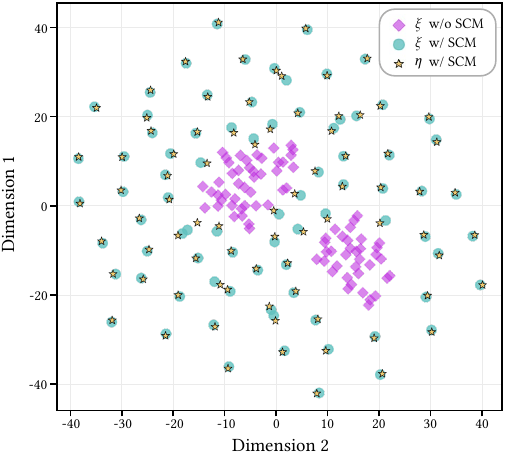}
  \caption{t-SNE visualization comparison of the key latent encodings with and without our proposed stage of SCM.
  }
  \label{fig:fig6}
\end{figure}

To further demonstrate the effectiveness of modeling cross-domain consistency in our CDCRec, we present qualitative performance in the stage of SCM across different datasets in Table ~\ref{tab:table2}. Sparse visibility inherently contains limited information, while the image is spatially dense and highly informative, making the alignment of two domains challenging. CDCRec delivers consistently high contrastive accuracy across all datasets, averaging 99.93\%. Different loss terms successfully converge to a notably low value, showcasing the effectiveness of our design. These performance results suggest that the emphasis on the multimodal alignment in the stage of SCM is essential, providing a solid foundation for interferometric data reconstruction.

\subsection{Qualitative Analysis}
We visualize key latent encodings $\xi$ and $\eta$ with and without the stage of SCM by utilizing the t-distributed Stochastic Neighbor Embedding (t-SNE) \cite{maaten2008visualizing}, where strong dependencies are exhibited. With the aid of the stage of SCM, strong connections are established, as it is demonstrated that each pair of the two modalities $\left\{ \xi, \eta \right\}$ is well matched. As visualized in the feature space, $\xi$ obtained without the stage of SCM is significantly more constrained, covering a much smaller area than those generated by our CDCRec. The broader distribution demonstrates that our proposed approach learns more diverse and comprehensive feature patterns, showcasing that explicit modeling of the cross-domain consistency is essential for successfully extracting the dependencies between two domains.

\begin{table}[t]
    \centering
    \caption{Qualitative results across different datasets in the stage of SCM. The contrastive accuracy ($Acc\uparrow$: higher is better) and different loss terms$\downarrow$ (lower is better) are reported.}
\resizebox{\columnwidth}{!}{
    \begin{tabular}{ll|cccc}
    \toprule
\multicolumn{2}{c|}{\textbf{Metric}}     & \textbf{MG} & \textbf{IRSG} & \textbf{UTSG} & \textbf{EGB} \\ \cmidrule{1-6}
\greenrow Contrastive Accuracy & $Acc\uparrow$                      & 100.00\%    & 99.87\%          & 99.86\%           & 100.00\%        \\
Contrastive Loss &  $\mathcal{L}^c\downarrow$                 & 0.0055          & 0.0246           & 0.0321           & 0.0073      \\ \cmidrule{1-6}
Image Recon. Loss &  $\mathcal{L}^{rec}_i\downarrow$         & 0.0489          & 0.1081           & 0.0673           & 0.0660      \\
Visibility Recon. Loss &  $\mathcal{L}^{rec}_v\downarrow$    & 0.0166          & 0.0116           & 0.0113           & 0.0118      \\ \cmidrule{1-6}
Total Loss &  $\mathcal{L}^{SCM}\downarrow$                        & 0.0658          & 0.1209           & 0.0802           & 0.0782      \\
    \bottomrule
    \end{tabular}
    }
    \label{tab:table2}
\end{table}

\begin{table}[t]
    \centering
    \caption{Comparisons of results between cross-dataset generalizations and intra-dataset performances.}
\resizebox{\columnwidth}{!}{
    \begin{tabular}{lc|cccc}
    \toprule
\textbf{Metric}          &\textbf{Settings in stage of SCM}              & \textbf{MG} & \textbf{IRSG} & \textbf{UTSG} & \textbf{EGB} \\ \cmidrule{1-6}
\multirow{2}{*}{PSNR$\uparrow$}          & Cross-Dataset          & 26.506          & 28.081           & 27.497           & 26.989          \\
          & Intra-Dataset          & 26.112           & 28.194           & 27.438           & 26.951          \\ \cmidrule{1-6}
\multirow{2}{*}{SSIM$\uparrow$}          & Cross-Dataset          & 0.914           & 0.925           & 0.924           & 0.917          \\
          & Intra-Dataset          & 0.912           & 0.926           & 0.924           & 0.918          \\
    \bottomrule
    \end{tabular}
    }
    \label{tab:table3}
\end{table}

\subsection{Cross-Dataset Generalization}
In this experiment, we employ two distinct configurations in the stage of SCM. The first configuration is cross-dataset generalization. For each specific dataset, we utilize the other three datasets as the training source in the stage of SCM. The second one is intra-dataset configuration. The entire training process is conducted exclusively on each individual dataset, respectively. As shown in Table ~\ref{tab:table3}, both configurations yield comparable results across datasets, indicating that the cross-domain consistency learned by our CDCRec demonstrates good generalization capabilities. Even though the cross-domain consistency is learned from other datasets, it is validated to be still effective in facilitating interferometric data reconstruction.

\begin{figure*}[t]
  \centering
  \includegraphics[width=\linewidth]{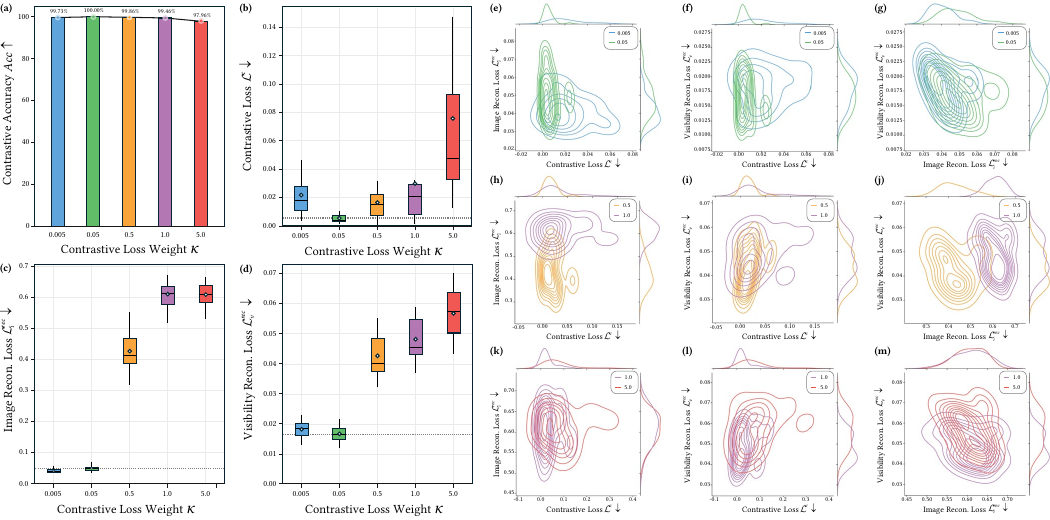}
  \caption{Illustration of the results of the sensitivity analysis. (a) to (d) show the distributions of contrastive accuracy $Acc$, contrastive loss $\mathcal{L}^c$, image reconstruction loss $\mathcal{L}^{rec}_i$, and visibility reconstruction loss $\mathcal{L}^{rec}_v$ when hyperparameter contrastive loss weight $\kappa$ changes. (e) to further demonstrate the bivariate joint kernel density plot of each pair of losses.}
  \label{fig:fig7}
\end{figure*}

\begin{table}[t]
    \centering
    \caption{Ablation study of stages and modules in CDCRec. Modules in the stage of SCM are grouped by pretext tasks to ensure the model remains functional.}
\resizebox{\columnwidth}{!}{
    \begin{tabular}{l|ccccc}
    \toprule
\multirow{2}{*}[-0.5ex]{\textbf{Model}}          &\multicolumn{2}{c}{\textbf{Stage of SCM}}              & \multirow{2}{*}[-0.5ex]{\textbf{Stage of IDR}}   &\multicolumn{2}{c}{\textbf{MG}} \\ \cmidrule(lr){2-3} \cmidrule(lr){5-6}
        &Task 1         &Task 2     &        & PSNR$\uparrow$        & SSIM$\uparrow$ \\ \cmidrule{1-6}
\Rmnum{1}          & \graycross          & \ding{51}          & \ding{51}           & 23.085           & 0.898  \\
\Rmnum{2}          & \ding{51}          & \graycross          & \ding{51}           & 23.005           & 0.901  \\
\Rmnum{3}          & \graycross          & \graycross          & \ding{51}           & 22.274           & 0.888  \\
\Rmnum{4}          & \ding{51}          & \ding{51}          & \graycross           & 12.286           & 0.719  \\  \cmidrule{1-6}
Full          & \ding{51}          & \ding{51}          & \ding{51}           & 26.506           & 0.914  \\ 
    \bottomrule
    \end{tabular}
    }
    \label{tab:table4}
\end{table}

\subsection{Sensitivity and Ablation Experiments}
We investigate the sensitivity of our CDCRec to its hyperparameter $\kappa$, as shown in Figure ~\ref{fig:fig7}. Figure ~\ref{fig:fig7} (a) to (d) show the distributions of contrastive accuracy $Acc$, contrastive loss $\mathcal{L}^c$, image reconstruction loss $\mathcal{L}^{rec}_i$, and visibility reconstruction loss $\mathcal{L}^{rec}_v$ when $\kappa$ changes by orders of magnitude. Our model exhibits stable performance with some increase in losses under extreme conditions, demonstrating its robustness. Furthermore, Figure ~\ref{fig:fig7} (e) to (m) demonstrate the bivariate joint kernel density plot of each pair of losses. With the gradual increase of $\kappa$, $\mathcal{L}^{rec}_i$ and $\mathcal{L}^{rec}_v$ exhibit an earlier upward trend and display notable synchronization in their distributional patterns. This phenomenon underscores the strong interplay between the two domains. Meanwhile, $\mathcal{L}^c$ receives greater emphasis and thus maintains a stably low level during the early stages. Ultimately, under extreme conditions, $\mathcal{L}^c$ exhibits a sudden surge alongside $\mathcal{L}^{rec}_i$ and $\mathcal{L}^{rec}_v$, which clearly demonstrates the collaborative effects among the three losses.

To systematically evaluate our CDCRec, we categorize its components based on two stages and pretext tasks, and conduct ablation experiments, as shown in Table ~\ref{tab:table4}. The performance of Model \Rmnum{1} and \Rmnum{2} demonstrates that removing specific pretext tasks leads to noticeable performance degradation. After removing the entire stage of SCM, Model \Rmnum{3} fails to explicitly model the cross-domain consistency, making its performance drop to the level of previous unimodal methods. Similarly, Model \Rmnum{4} is the variant without the stage of IDR. Its performance decrease suggests that the learned prior knowledge in the stage SCM has to be transferred to the task of interferometric data reconstruction to take effect, highlighting the importance of the stage of IDR.

\section{Conclusion and Future Work}
In this paper, we present a multimodal radio interferometric data reconstruction method that explicitly models cross-domain consistency, called CDCRec. It can uncover the underlying dependencies and correlations between the visibility and image domains in a multimodal manner with a well-structured hierarchical multi-task and multi-stage framework to tackle the current challenges in interferometric data reconstruction: (1) the highly ill-posed nature of reconstructing the image from sparse visibility, (2) the intrinsic difficulty in the modeling of strong dependencies between the sparse visibility and image. Extensive experimental results demonstrate competitive performance across multiple datasets, exhibiting robustness and generalization of CDCRec as well. Our work underscores the importance of cross-domain consistency for interferometric data reconstruction and offers foundations and insights for multimodal modeling. We envision that the cross-domain consistency could benefit other disciplines involving domain transformations with increased diversity in both numerical scales and heterogeneous types. Future work could also involve exploring fields such as magnetic resonance imaging and remote sensing to demonstrate the versatility of our approach.

\bibliographystyle{ACM-Reference-Format}
\bibliography{main}

\end{document}